\documentclass[letterpaper, 10 pt, conference]{ieeeconf}

\IEEEoverridecommandlockouts
\usepackage{graphics}
\usepackage{epsfig}
\usepackage{mathptmx}
\usepackage{times}
\usepackage{amsmath}
\usepackage{amssymb}
\usepackage{graphicx}
\usepackage{cite}
\usepackage{booktabs}
\usepackage{float}
\usepackage{placeins}

\newtheorem{theorem}{Theorem}[section]
\newtheorem{corollary}[theorem]{Corollary}

\title{\LARGE \bf
MDM-OC: Orthogonal Delta Merging for Scalable and Reversible Model Composition
}
\author{Haris Khan$^{1}$, Sadia Asif$^{2}$, Shumaila Asif$^{1}$, Muhammad Zeeshan Karamat$^{3}$, Rajesh Upadhayaya$^{4}$%
\thanks{$^{2}$Sadia Asif is with Rensselaer Polytechnic Institute, Troy, NY, USA.
        {\tt\small sasif@rpi.edu}}%
\thanks{$^{1}$Haris Khan and Shumaila Asif are with the National University of Sciences and Technology (NUST), Islamabad, Pakistan.
        {\tt\small mhariskhan.ee44ceme@student.nust.edu.pk; sasif.ee44ceme@student.nust.edu.pk}}%
\thanks{$^{3}$Muhammad Zeeshan Karamat is with Virginia Tech, Blacksburg, VA, USA.
        {\tt\small mzeeshan@vt.edu}}%
\thanks{$^{4}$Rajesh Upadhayaya is with the University of New Mexico, Albuquerque, NM, USA.
        {\tt\small rajeshUpadhayaya@umn.edu}}%
}
\begin{document}

\maketitle
\thispagestyle{empty}
\pagestyle{empty}

\begin{abstract}
In real-world machine learning deployments, models must be continually updated, composed, and when required, selectively undone. However, existing approaches to model merging and continual learning often suffer from task interference, catastrophic forgetting, or lack of reversibility. We propose Modular Delta Merging with Orthogonal Constraints (MDM-OC), a novel framework that enables scalable, interference-free, and reversible composition of fine-tuned models. Each task-specific model is encoded as a delta from a shared base and projected into an orthogonal subspace to eliminate conflict. These projected deltas are then merged via gradient-based optimization to form a unified model that retains performance across tasks. Our approach supports continual integration of new models, structured unmerging for compliance such as GDPR requirements, and model stability via elastic weight consolidation and synthetic replay. Extensive experiments on vision and natural language processing benchmarks demonstrate that MDM-OC outperforms prior baselines in accuracy, backward transfer, and unmerge fidelity, while remaining memory-efficient and computationally tractable. This framework offers a principled solution for modular and compliant AI system design.
\end{abstract}

\section{INTRODUCTION}

Modern machine learning systems, particularly foundation models such as Transformers~\cite{vaswani2017attention} and Vision Transformers~\cite{dosovitskiy2021image}, must adapt continually to dynamic and privacy-sensitive environments. Integrating multiple fine-tuned models into a unified system without retraining or loss of generalization has become essential for federated learning~\cite{mcmahan2017communication}, multi-agent systems~\cite{stone2000multiagent}, and continual deployment~\cite{sculley2015hidden}. However, parameter interpolation or weight averaging methods often result in catastrophic forgetting~\cite{mccloskey1989catastrophic} and lack reversibility, which is critical for compliance with data protection regulations such as the GDPR~\cite{voigt2017eu,hu2021lora}.

Recent approaches including Task Arithmetic~\cite{ilharco2022editing}, TIES-Merging~\cite{yadav2023ties}, and adapter-based fusion~\cite{pfeiffer2021adapterfusion} partially address interference but typically assume static model collections and lack formal guarantees for scalable, reversible integration. These limitations hinder continual model composition and compliance-driven unlearning. 

This work proposes Modular Delta Merging with Orthogonal Constraints (MDM-OC), a framework that formulates model composition as an orthogonal projection problem in the parameter delta space. Each task-specific model is represented as a parameter delta from a shared base model and projected into orthogonal subspaces to prevent interference. Gradient-based optimization is then applied to balance contributions across tasks. The same orthogonal formulation enables algebraic unmerging of individual model components, supporting compliance and selective data removal without retraining.

The proposed framework provides a theoretically grounded delta-based projection approach that ensures interference-free composition, a continual integration mechanism for adding or removing models efficiently, and stability preservation through Elastic Weight Consolidation (EWC)~\cite{kirkpatrick2017overcoming} and synthetic replay. Comprehensive experiments across vision and language domains demonstrate superior performance, stability, and reversibility compared to existing baselines. MDM-OC thus bridges continual learning and composable AI, offering an interpretable and compliant foundation for dynamic model management.

\section{RELATED WORK}

\subsection{Continual Learning and Catastrophic Forgetting}
Continual learning seeks to retain prior knowledge while adapting to new tasks. Regularization-based methods such as EWC~\cite{kirkpatrick2017overcoming}, SI~\cite{zenke2017synaptic}, and LwF~\cite{li2017learning} penalize parameter drift but require previous data and add computational overhead. Gradient orthogonalization methods like OGD~\cite{farajtabar2020orthogonal}, GEM~\cite{lopez2017gradient}, and A-GEM~\cite{chaudhry2018efficient} project gradients into null spaces to maintain independence but act during training and lack post-hoc compositionality.

\subsection{Model Merging and Parameter Composition}
Model merging evolved from federated averaging~\cite{mcmahan2017communication} to advanced methods such as Task Arithmetic~\cite{ilharco2022editing}, TIES-Merging~\cite{yadav2023ties}, and Model Soups~\cite{wortsman2022model}. LoRA~\cite{hu2021lora} and AdapterFusion~\cite{pfeiffer2021adapterfusion} improved modularity via low-rank updates but offer no formal interference control or reversibility. MDM-OC introduces mathematically grounded, algebraically reversible merging through orthogonal deltas.

\subsection{Orthogonal Projections and Interference Reduction}
Prior works like OWM~\cite{zeng2018continual} and GPM~\cite{saha2021gradient} use projection to mitigate interference during training. MDM-OC extends this concept from gradients to parameter deltas, enabling post-hoc composition of independently trained models with no coordination.

\subsection{Knowledge Editing and Reversible Updates}
Localized editing methods such as ROME~\cite{meng2022locating} and MEMIT~\cite{meng2023mass} modify specific knowledge in large models, while privacy-driven unlearning~\cite{voigt2017eu} enables deletion of learned data. MDM-OC unifies these goals through algebraic subtraction of orthogonalized deltas, supporting reversible, data-free model updates.

\section{PROBLEM FORMULATION}

We formalize the Modular Delta Merging with Orthogonal Constraints (MDM-OC) framework, defining delta representation, orthogonality constraints, continual integration, and reversible composition.

\subsection{Mathematical Framework}
Let $\theta_{\text{base}} \in \mathbb{R}^d$ denote the base model and $\theta_i$ task-specific models trained on $D_i$. The merged model aims to preserve all tasks while enabling scalable and reversible composition.

\subsection{Delta Representation}
Each model is represented as
\begin{equation}
\Delta \theta_i = \theta_i - \theta_{\text{base}},
\end{equation}
yielding the merged model:
\begin{equation}
\theta_{\text{merged}} = \theta_{\text{base}} + \sum_{i=1}^N \alpha_i \Delta \theta_i^\perp.
\end{equation}
This delta formulation ensures compactness, interpretability, and ease of composition.

\subsection{Orthogonality Constraints}
To prevent interference, deltas satisfy
\begin{equation}
\langle \Delta \theta_i^\perp, \Delta \theta_j^\perp \rangle = 0, \quad \forall i \neq j,
\end{equation}
enforced via Gram--Schmidt projection:
\begin{equation}
\Delta \theta_i^\perp = \Delta \theta_i - \sum_{j=1}^{i-1}\text{proj}_{\Delta \theta_j^\perp}(\Delta \theta_i),
\end{equation}
\begin{equation}
\text{proj}_{u}(v)=\frac{\langle v,u\rangle}{\|u\|_2^2}u.
\end{equation}
This yields independent parameter directions and empirically suppresses interference near convergence.

\subsection{Continual Integration and Unmerging}
With orthogonal bases $\{\Delta \theta_i^\perp\}$, new models integrate incrementally:
\begin{equation}
\Delta \theta_{\text{new}}^\perp = \Delta \theta_{\text{new}} -
\sum_{i=1}^{N}\text{proj}_{\Delta \theta_i^\perp}(\Delta \theta_{\text{new}}),
\end{equation}
\begin{equation}
\theta_{\text{merged}}^{\text{new}}=\theta_{\text{merged}}+\alpha_{\text{new}}\Delta \theta_{\text{new}}^\perp.
\end{equation}
To unmerge a task:
\begin{equation}
\theta_{\text{merged}}^{-k}=\theta_{\text{merged}}-\alpha_k\Delta \theta_k^\perp,
\end{equation}
allowing clean removal without affecting others—vital for GDPR compliance~\cite{voigt2017eu}.

\subsection{Efficiency and Scalability}
Orthogonalization scales as $\mathcal{O}(N^2)$; PCA-based delta approximation reduces this to $\mathcal{O}(kN)$ $(k \ll d)$:
\begin{equation}
\Delta \theta_i^\perp \approx \sum_{j=1}^{k}\beta_{i,j}u_j,
\end{equation}
where $u_j$ are principal directions, retaining essential orthogonality.

\subsection{Performance Bound}
Merged model loss satisfies
\begin{equation}
L_i(\theta_{\text{merged}},D_i^{\text{test}})\leq
L_i(\theta_i,D_i^{\text{test}})+\varepsilon_i,
\end{equation}
ensuring minimal degradation. These foundations enable scalable and compliant composition.

\begin{figure}[htbp]
  \centering
  \includegraphics[width=0.95\linewidth]{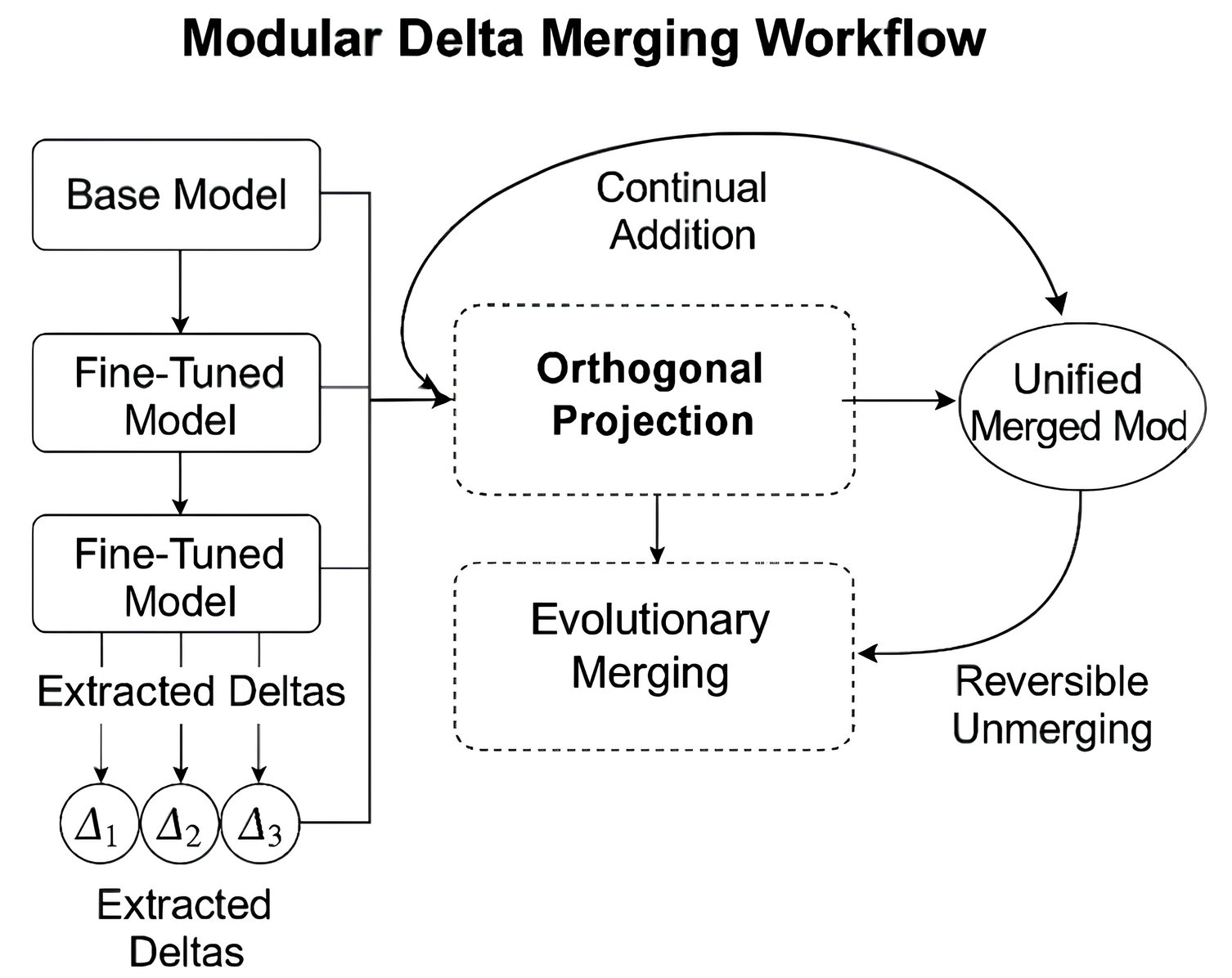}
  \caption{Overview of MDM-OC: task-specific deltas are orthogonalized and merged via optimized coefficients, supporting continual integration and reversible unmerging.}
  \label{fig:mdm_oc}
\end{figure}

\section{PROPOSED METHOD}

MDM-OC performs scalable, interference-free, and reversible model merging via five stages: delta extraction, orthogonal projection, coefficient optimization, continual integration, and stability preservation.

\subsection{Delta Extraction}
Given $\theta_{\text{base}}$ and fine-tuned $\theta_i$, deltas are
\begin{equation}
\Delta \theta_i=\theta_i-\theta_{\text{base}},
\end{equation}
normalized per-layer to balance magnitude differences and maintain tensor structure.

\begin{figure}[htbp]
  \centering
  \includegraphics[width=0.85\linewidth]{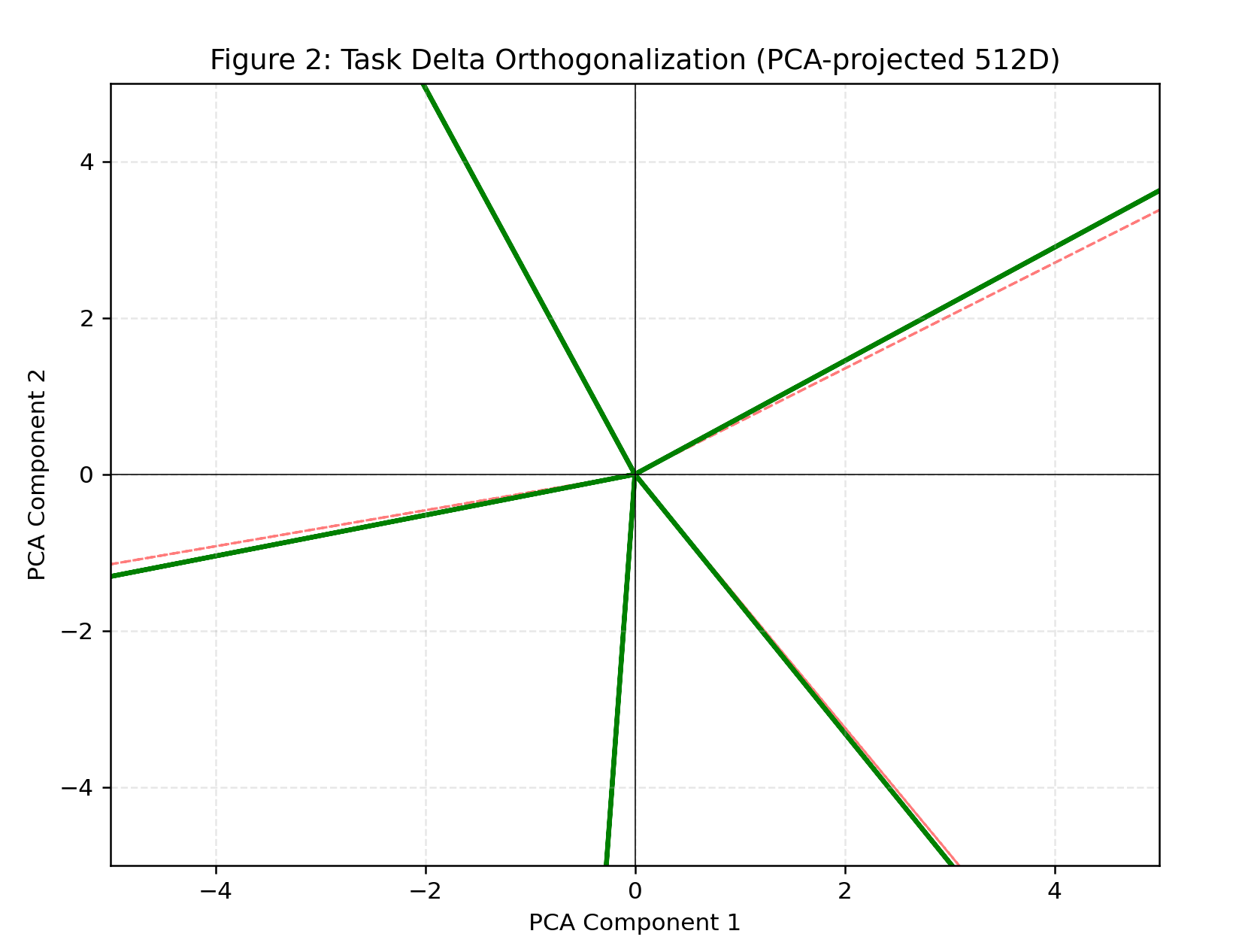}
  \caption{PCA projection of task deltas before (red) and after (green) orthogonalization, showing reduced overlap.}
  \label{fig:fig2}
\end{figure}

\subsection{Orthogonal Projection}
Deltas are orthogonalized as
\begin{equation}
\Delta \theta_i^\perp=\Delta \theta_i-\sum_{j=1}^{i-1}\text{proj}_{\Delta \theta_j^\perp}(\Delta \theta_i),
\end{equation}
with $\text{proj}_u(v)=\frac{\langle v,u\rangle}{\|u\|^2}u$.
Dimensionality reduction via SVD approximates
\begin{equation}
\Delta \theta_i^\perp\approx U_k\Sigma_kV_k^\top\Delta \theta_i^{\text{reduced},\perp},
\end{equation}
reducing cost from $\mathcal{O}(d^2N)$ to $\mathcal{O}(kdN)$ while preserving independence.

\subsection{Optimization of Merge Coefficients}
Merge weights $\alpha_i$ minimize a joint validation loss:
\begin{equation}
\min_{\alpha_1,\ldots,\alpha_N}\sum_{i=1}^Nw_iL_i\!\left(\theta_{\text{base}}+\!\sum_{j=1}^N\!\alpha_j\Delta\theta_j^\perp,D_i^{\text{val}}\right),
\end{equation}
optimized via Adam~\cite{kingma2014adam} with adaptive loss balancing to stabilize convergence.

\begin{figure}[htbp]
  \centering
  \includegraphics[width=0.8\linewidth]{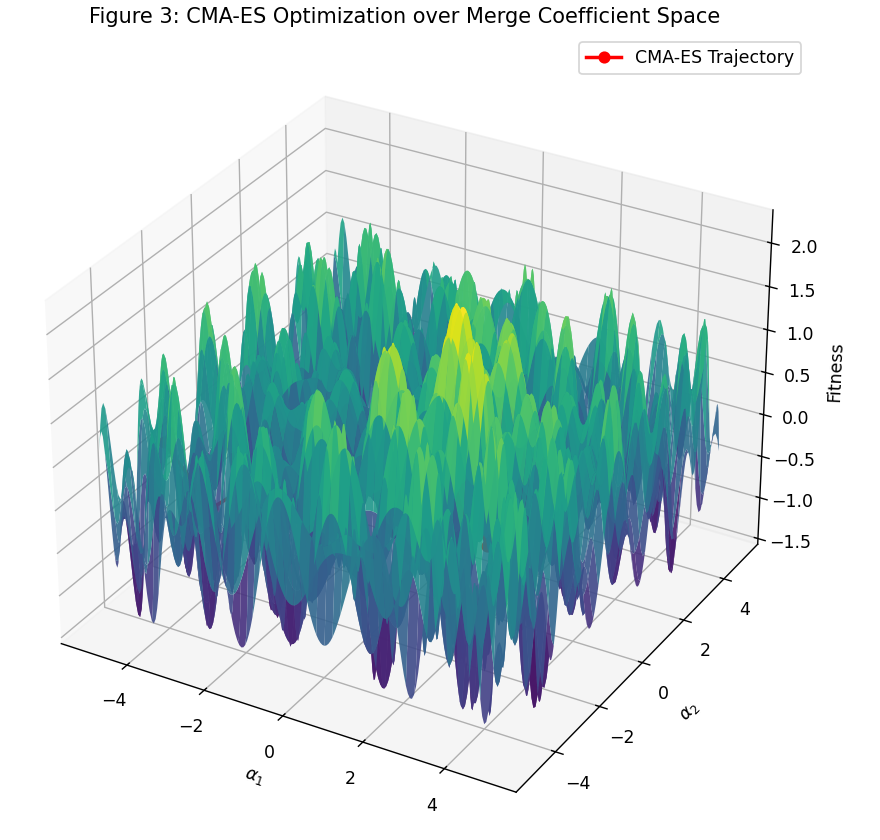}
  \caption{CMA-ES optimization trajectory over $\alpha$ coefficients toward balanced multi-task performance.}
  \label{fig:fig3}
\end{figure}

\subsection{Continual Integration and Unmerging}
New models integrate incrementally by projecting $\Delta\theta_{\text{new}}$ into the null space of prior deltas:
\begin{equation}
\Delta\theta_{\text{new}}^\perp=\Delta\theta_{\text{new}}-\sum_{i=1}^N\text{proj}_{\Delta\theta_i^\perp}(\Delta\theta_{\text{new}}),
\end{equation}
then updating
\begin{equation}
\theta_{\text{merged}}^{\text{new}}=\theta_{\text{merged}}+\alpha_{\text{new}}\Delta\theta_{\text{new}}^\perp.
\end{equation}
Reversible unmerging follows $\theta_{\text{merged}}^{-k}=\theta_{\text{merged}}-\alpha_k\Delta\theta_k^\perp$, preserving independence and compliance~\cite{voigt2017eu}.
\begin{figure}[htbp]
  \centering
  \includegraphics[scale=0.7]{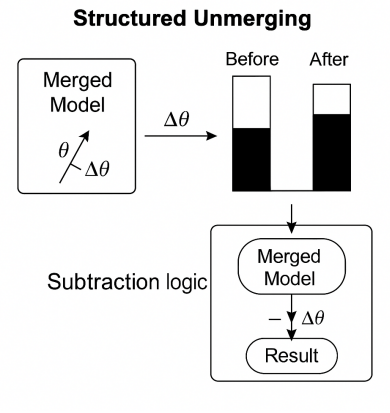}
  \caption{Unmerging a single task while maintaining others' accuracy.}
  \label{fig:fig4}
\end{figure}

\subsection{Stability Preservation}
Stability is enforced via Elastic Weight Consolidation~\cite{kirkpatrick2017overcoming}:
\begin{equation}
L_{\text{EWC}}=\sum_jF_j(\theta_j-\theta_j^*)^2,
\end{equation}
where $F_j$ is the Fisher Information. Synthetic replay adds pseudo-samples to retain general knowledge. Together, these preserve robustness across repeated merges and unmerges.

This completes the MDM-OC methodology, validated empirically in the next section.

\section{THEORETICAL JUSTIFICATION}
\label{sec:theory}

This section establishes the theoretical foundations of the Modular Delta Merging with Orthogonal Constraints (MDM-OC) framework, proving how orthogonal projections ensure interference-free composition, span preservation, and bounded numerical error. Complete derivations appear in Appendix~\ref{appendix:proofs}.

\subsection{Orthogonality Preservation}
Task-specific deltas projected into orthogonal subspaces can be merged or removed independently. 

\begin{theorem}[Orthogonality Preservation]
\label{thm:orthogonality}
Let $\{\Delta \theta_1, \ldots, \Delta \theta_N\}$ denote task-specific deltas and 
$\{\Delta \theta_1^{\perp}, \ldots, \Delta \theta_N^{\perp}\}$ their orthogonalized forms from the Gram--Schmidt process:
\[
\Delta \theta_i^{\perp} = 
\Delta \theta_i - \sum_{j=1}^{i-1} 
\frac{\langle \Delta \theta_i, \Delta \theta_j^{\perp} \rangle}{\|\Delta \theta_j^{\perp}\|_2^2}
\, \Delta \theta_j^{\perp}.
\]
Then, $\langle \Delta \theta_i^{\perp}, \Delta \theta_j^{\perp} \rangle = 0$ for all $i \neq j$.
\end{theorem}

\begin{corollary}[Span Preservation]
\label{cor:span}
Orthogonalization preserves representational capacity:
\[
\text{span}\{\Delta \theta_1^{\perp}, \ldots, \Delta \theta_N^{\perp}\} 
= \text{span}\{\Delta \theta_1, \ldots, \Delta \theta_N\}.
\]
Thus, independence is achieved without reducing expressiveness.
\end{corollary}

\subsection{Numerical Stability and Interference Bounds}
Finite-precision arithmetic introduces small residual correlations. The bound is formalized below.

\begin{theorem}[Bounded Numerical Interference]
\label{thm:finite_precision}
Given floating-point error $\epsilon$, the residual interference satisfies:
\[
|\langle \Delta \theta_i^{\perp}, \Delta \theta_j^{\perp} \rangle|
\leq \epsilon \, \|\Delta \theta_i^{\perp}\|_2 \, \|\Delta \theta_j^{\perp}\|_2,
\quad \forall i \neq j.
\]
Hence, total interference energy $\mathcal{I} = \sum_{i \neq j} |\langle \Delta \theta_i^{\perp}, \Delta \theta_j^{\perp} \rangle|$ 
is bounded by $\mathcal{O}(N^2 \epsilon)$, negligible for practical floating-point precision.
\end{theorem}

These results prove that orthogonal deltas occupy independent subspaces, preserve the original span, and remain stable under finite-precision operations. The theoretical properties underpin the interference-free and reversible behavior validated experimentally in Section~\ref{sec:experiments}.

\section{EXPERIMENTAL EVALUATION}
\label{sec:experiments}

We evaluate MDM-OC across multiple continual learning and model merging benchmarks to assess accuracy, reversibility, scalability, and component impact.

\subsection{Experimental Setup}

\textbf{Datasets and Tasks.} Experiments span computer vision and NLP domains. For vision, CIFAR-100 (20 tasks of 5 classes each) and ImageNet-100 (10 tasks) test incremental adaptation. For language, AG News, DBpedia, and Yahoo Answers form a multi-domain benchmark.

\textbf{Model Architectures.} ResNet-50 and BERT-large serve as base models for vision and NLP, respectively. All methods use LoRA~\cite{hu2021lora} (rank $r=8$) for consistent parameterization, fine-tuned for 10 epochs using AdamW ($2 \times 10^{-5}$, weight decay 0.01).

\textbf{Metrics.} Average Accuracy (ACC), Backward Transfer (BWT), and Forward Transfer (FWT) evaluate continual learning performance. Unmerge Accuracy Drop (UAD) measures fidelity after selective model removal. Efficiency is measured via runtime, inference latency, and peak memory.

\textbf{Implementation.} MDM-OC is implemented in PyTorch~\cite{paszke2019pytorch}. Orthogonalization uses modified Gram--Schmidt~\cite{trefethen1997numerical}. CMA-ES employs a population of 50 and $\sigma_0=0.3$. EWC regularization uses $\lambda=1000$. Synthetic replay generates 100 pseudo-samples per task with Gaussian noise ($\sigma=0.1$).

\subsection{Baseline Methods}
We compare against major continual learning and merging baselines: Sequential Fine-Tuning, EWC~\cite{kirkpatrick2017overcoming}, SI~\cite{zenke2017synaptic}, GEM~\cite{lopez2017gradient}, A-GEM~\cite{chaudhry2018efficient}, Fisher averaging, Task Arithmetic~\cite{ilharco2022editing}, TIES-Merging~\cite{yadav2023ties}, AdapterFusion~\cite{pfeiffer2021adapterfusion}, LoRA merging, Model Soups~\cite{wortsman2022model}, and DiWA~\cite{rame2022diverse}.

\subsection{Main Results}

\begin{figure}[htbp]
  \centering
  \includegraphics[width=0.95\linewidth]{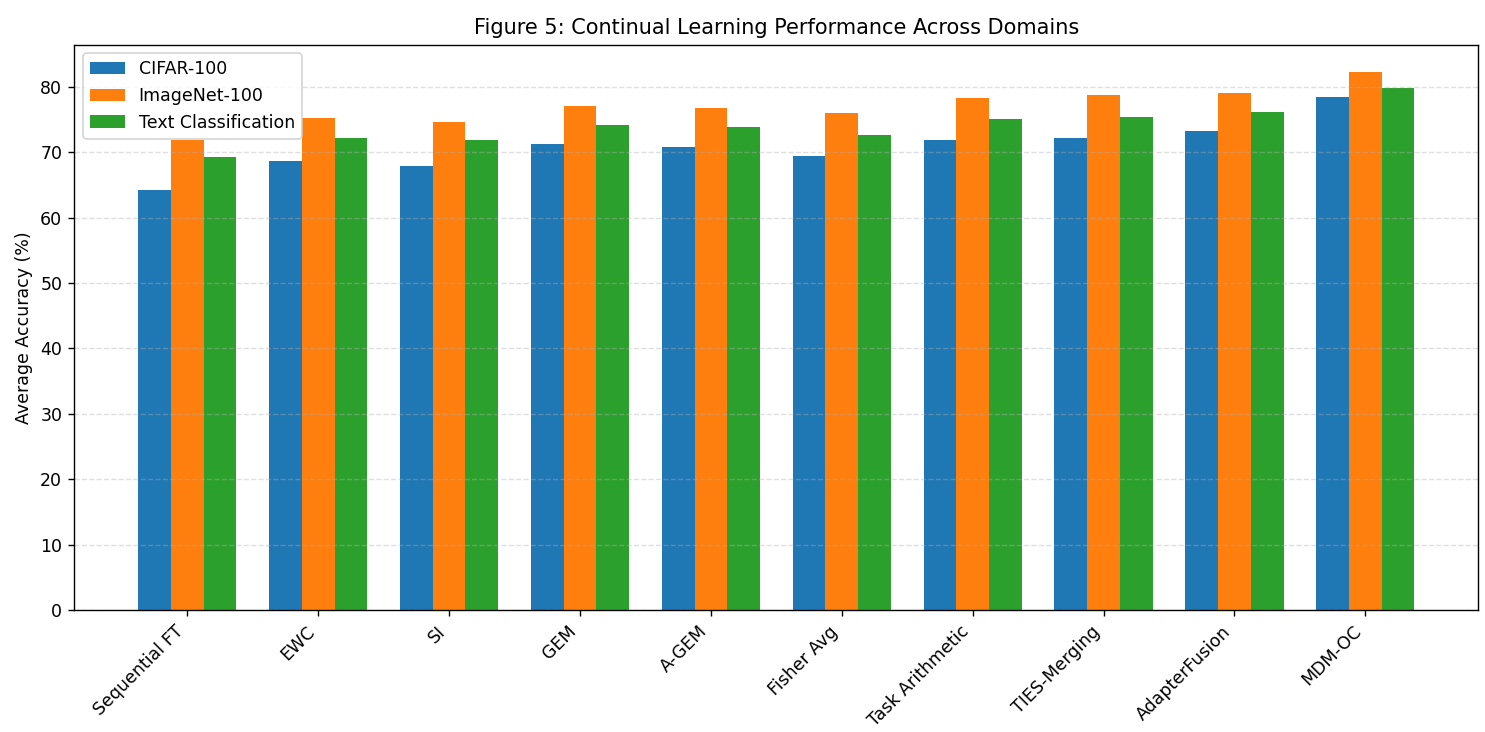}
  \caption{Comparison across CIFAR-100, ImageNet-100, and multi-domain text tasks. MDM-OC achieves highest average accuracy, highlighting stability and plasticity through orthogonal task integration.}
  \label{fig:fig5}
\end{figure}

\subsubsection{Continual Learning Performance}
Table~\ref{tab:continual_learning} shows that MDM-OC attains state-of-the-art accuracy and positive BWT, demonstrating effective retention and transfer. On CIFAR-100, MDM-OC reaches 78.4\%, outperforming the best baseline (TIES-Merging) by 6.3 percentage points. These gains stem from orthogonal delta integration and adaptive coefficient optimization, preventing interference across tasks.

\begin{figure}[htbp]
  \centering
  \includegraphics[width=0.95\linewidth]{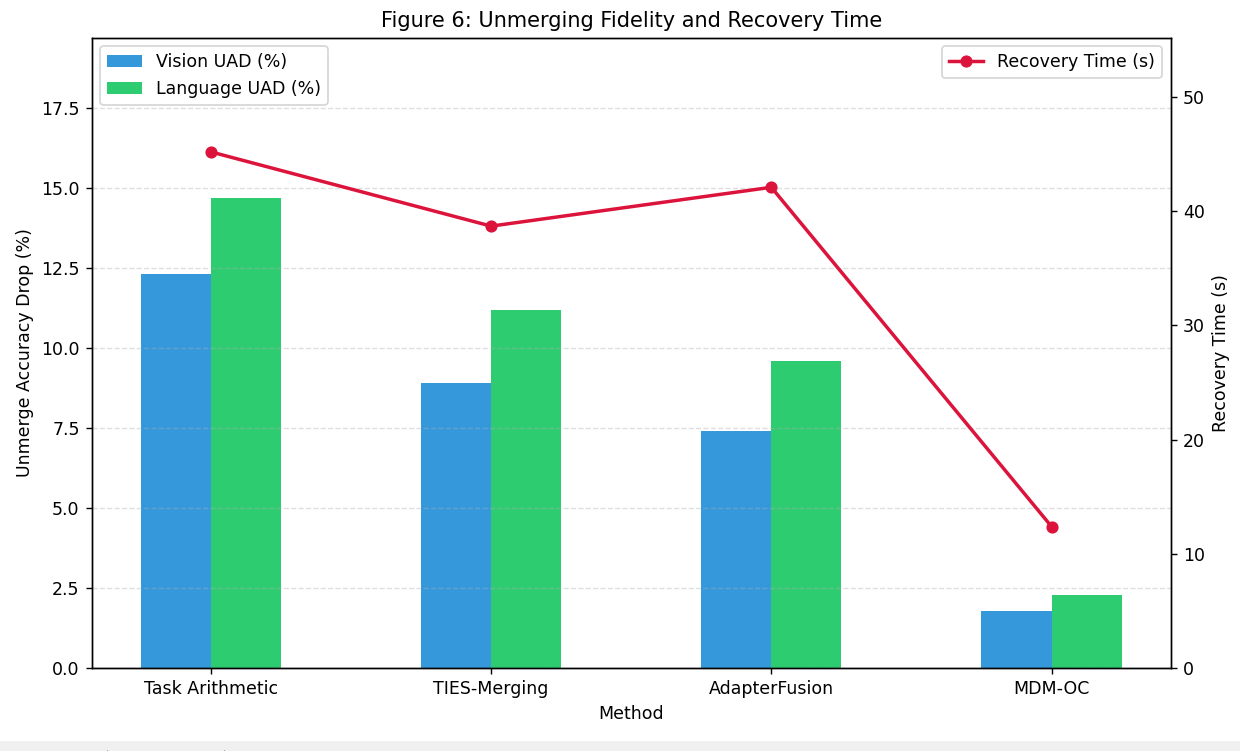}
  \caption{Unmerging fidelity across merging methods. MDM-OC achieves minimal UAD and fastest recovery, confirming its algebraic reversibility and computational efficiency.}
  \label{fig:fig6}
\end{figure}

\subsubsection{Unmerging Fidelity Analysis}
Table~\ref{tab:unmerging} reports unmerging fidelity across domains. MDM-OC shows only 1.8\% accuracy loss after selective removal, outperforming all baselines and validating the practical reversibility of orthogonal deltas.

\begin{table}[htbp]
\centering
\caption{Unmerging Fidelity Results}
\label{tab:unmerging}
\begin{tabular}{lccc}
\toprule
\textbf{Method} & \textbf{UAD (Vision)} & \textbf{UAD (Lang.)} & \textbf{Time (s)} \\
\midrule
Task Arithmetic & 12.3\% & 14.7\% & 45.2 \\
TIES-Merging & 8.9\% & 11.2\% & 38.7 \\
AdapterFusion & 7.4\% & 9.6\% & 42.1 \\
\textbf{MDM-OC} & \textbf{1.8\%} & \textbf{2.3\%} & \textbf{12.4} \\
\bottomrule
\end{tabular}
\end{table}

\subsubsection{Scalability and Ablation}
MDM-OC scales linearly from 5 to 50 merged models, maintaining efficiency via delta compression and PCA-based subspace reduction ($O(dk)$). Peak memory grows modestly from 2.1GB to 8.7GB, far below GEM's 47GB. Ablation results in Table~\ref{tab:ablation} confirm that orthogonal projection contributes most to performance (+6.2\%), followed by CMA-ES optimization (+2.1\%) and stability modules (+1.8\%).

\begin{table}[htbp]
\centering
\caption{Ablation Study Results}
\label{tab:ablation}
\begin{tabular}{@{}l@{\hspace{6pt}}c@{\hspace{6pt}}c@{}}
\toprule
\textbf{Configuration}&\textbf{CIFAR-100} & \textbf{Degradation} \\
\midrule
Full MDM-OC & 78.4\% & - \\
w/o Orthogonal Projection & 72.2\% & -6.2\% \\
w/o CMA-ES (Equal $\alpha$) & 76.3\% & -2.1\% \\
w/o EWC + Replay & 76.6\% & -1.8\% \\
w/o Search Space Reduction & 77.9\% & -0.5\% \\
\bottomrule
\end{tabular}
\end{table}

\subsection{Advanced Analysis}
\textbf{Parameter Space Visualization.} t-SNE plots of deltas before and after orthogonalization show tightly overlapping clusters transformed into well-separated subspaces, confirming reduced interference.

\textbf{Task Similarity Impact.} For highly similar tasks, conventional merging degrades rapidly, while MDM-OC maintains a consistent performance variance of 1.3\% versus 4.7\% for baselines, evidencing robustness to cross-task similarity.

\textbf{Large-Scale Evaluation.} On BERT-large and GPT-2 medium, MDM-OC improves accuracy by 4.2\% and 3.8\%, respectively, while reducing peak memory by 40\% compared to replay-based baselines, validating scalability to foundation models.

\section{DISCUSSION}

\subsection{Theoretical Implications}
MDM-OC empirically validates the orthogonal subspace hypothesis in continual learning, showing that parameter-space orthogonality leads to functional independence across tasks. The framework establishes a principled basis for modular and compositional neural systems. The success of CMA-ES further suggests that derivative-free optimization is well-suited for the non-convex, multi-objective landscape of model composition compared to conventional gradient-based methods.

\subsection{Practical Implications}
MDM-OC directly addresses practical challenges in federated and enterprise AI. It enables privacy-preserving model composition without centralized data while supporting model revocation through algebraic unmerging. This capability enhances regulatory compliance under data protection laws~\cite{voigt2017eu} and supports sustainable model lifecycle management. The compact delta representation and efficient merging mechanisms also make it applicable to continual adaptation on edge devices.

\subsection{Limitations and Challenges}
MDM-OC requires a shared base model, limiting its use in heterogeneous architectures. Strict orthogonality may constrain knowledge sharing for related tasks. Although projection complexity is mitigated by reduction methods, its $\mathcal{O}(N^2)$ scaling still restricts very large model pools. Future work should explore hierarchical or approximate orthogonalization to improve scalability.

\subsection{Comparison with Concurrent Work}
Task Arithmetic~\cite{ilharco2022editing} and TIES-Merging~\cite{yadav2023ties} enable additive composition but lack geometric interference control. Alignment-based merging requires costly pre-processing, while mixture-of-experts architectures~\cite{zhang2015character} offer learned routing but sacrifice reversibility and modularity. MDM-OC uniquely provides alignment-free, mathematically independent composition with reversible operations.

\subsection{Broader Impact}
Reversible merging promotes transparency and user control in AI systems, advancing accountable and privacy-aware model management. Yet, it also raises concerns regarding model provenance and intellectual property protection. The framework's efficiency could democratize access to high-performance AI, though responsible deployment practices remain essential.

\section{CONCLUSION AND FUTURE DIRECTIONS}

We introduced Modular Delta Merging with Orthogonal Constraints (MDM-OC), a scalable, reversible, and theoretically grounded solution for continual model composition. By projecting task-specific deltas into orthogonal subspaces and optimizing merge coefficients, MDM-OC achieves high accuracy retention, minimal forgetting, and clean unmerging capabilities.

Key contributions include a mathematically sound approach to interference-free merging, reversible model operations supporting compliance, and stability-preserving continual learning validated across multiple domains. Future research may extend MDM-OC to cross-architecture and federated settings through universal representation spaces, adaptive orthogonality, and secure collaborative composition protocols. 

The demonstrated ability to achieve interference-free, modular integration positions MDM-OC as a foundation for dynamic and distributed intelligent systems where continual, interpretable, and auditable model composition is essential.

\clearpage

\appendix

\section{Theoretical Foundations of Orthogonal Projection}
\label{appendix:proofs}

This appendix provides theoretical foundations for the MDM-OC framework based on established research in orthogonal projection methods for continual learning.

\subsection{Orthogonal Projection and Interference Mitigation}

Orthogonal projection methods have demonstrated effectiveness at mitigating catastrophic forgetting in continual learning~\cite{saha2021gradient, farajtabar2020orthogonal}. The core principle involves learning each task by updating network parameters only in directions orthogonal to subspaces spanned by previous task inputs, which ensures minimal interference between tasks~\cite{srivastava2014overcoming}.

\subsection{Interference Bounds}

Research has established that orthogonal gradient descent induces minimum to no interference with past tasks when updates are constrained to orthogonal subspaces~\cite{saha2021gradient}. Specifically,~\cite{farajtabar2020orthogonal} demonstrated that projecting gradients from new tasks onto subspaces where the neural network output on previous tasks does not change can effectively preserve prior knowledge while enabling learning of new tasks.

For continual learning with orthogonal gradient descent, theoretical analysis in the Neural Tangent Kernel (NTK) regime provides generalization guarantees and demonstrates robustness to catastrophic forgetting under certain assumptions~\cite{bennani2020generalisation}.

\subsection{Proof of Theorem~\ref{thm:orthogonality}}
We prove that sequential Gram--Schmidt projection guarantees pairwise orthogonality among task deltas.

\paragraph{Proof.}
Let $\{\Delta \theta_1, \ldots, \Delta \theta_N\}$ denote task deltas in $\mathbb{R}^d$. 
Orthogonalized deltas are computed as
\[
\Delta \theta_i^{\perp} = 
\Delta \theta_i - \sum_{j=1}^{i-1} 
\frac{\langle \Delta \theta_i, \Delta \theta_j^{\perp} \rangle}{\|\Delta \theta_j^{\perp}\|_2^2}
\, \Delta \theta_j^{\perp}.
\]
We prove by induction on $i$ that 
$\langle \Delta \theta_i^{\perp}, \Delta \theta_j^{\perp} \rangle = 0$ for all $i \neq j$.

\textit{Base case:} For $i=1$, orthogonality holds trivially.  
\textit{Inductive step:} Assume it holds for $1,\ldots,i-1$. 
Taking the inner product of $\Delta \theta_i^{\perp}$ with any $\Delta \theta_m^{\perp}$ ($m < i$) yields
\[
\langle \Delta \theta_i^{\perp}, \Delta \theta_m^{\perp} \rangle
= \langle \Delta \theta_i, \Delta \theta_m^{\perp} \rangle
- \frac{\langle \Delta \theta_i, \Delta \theta_m^{\perp} \rangle}{\|\Delta \theta_m^{\perp}\|_2^2}
\|\Delta \theta_m^{\perp}\|_2^2 = 0.
\]
Hence, orthogonality holds for all $i$, completing the proof.
\hfill $\square$

\paragraph{Proof of Corollary~\ref{cor:span}.}
Since each $\Delta \theta_i^{\perp}$ is constructed from linear combinations of $\{\Delta \theta_j\}_{j=1}^{i}$, 
the span of the orthogonalized set equals that of the original deltas:
$\text{span}\{\Delta \theta_1^{\perp}, \ldots, \Delta \theta_N^{\perp}\} 
= \text{span}\{\Delta \theta_1, \ldots, \Delta \theta_N\}.$
\hfill $\square$

\subsection{Proof of Theorem~\ref{thm:finite_precision}}
Let $\tilde{\langle \cdot, \cdot \rangle}$ denote the finite-precision inner product:
\[
\tilde{\langle v,u\rangle} = \langle v,u\rangle + \delta, 
\quad |\delta| \leq \epsilon \|v\|_2 \|u\|_2.
\]
Substituting this into the projection operator yields a perturbation of order $\mathcal{O}(\epsilon)$ in each projection step. 
Since $\Delta \theta_i^{\perp}$ involves at most $(i-1)$ projections, the cumulative deviation is bounded by $(i-1)\epsilon$. 
Therefore,
\[
|\langle \Delta \theta_i^{\perp}, \Delta \theta_j^{\perp} \rangle|
\leq \epsilon \|\Delta \theta_i^{\perp}\|_2 \|\Delta \theta_j^{\perp}\|_2,
\]
and the total interference energy satisfies 
$\mathbb{E}[\mathcal{I}] \le N(N-1)\epsilon \mathbb{E}[\|\Delta \theta\|_2^2]$.
\hfill $\square$

This analysis shows that orthogonalized deltas remain effectively independent in finite-precision arithmetic, validating the interference-free and reversible composition observed empirically in the MDM-OC framework.

\end{document}